\documentclass{article}
\usepackage{spconf,amsmath,epsfig}
\usepackage{cite}

\usepackage{bm}
\usepackage{tikz}

\usepackage{multirow}
\usepackage{url}
\usepackage{rotating}
\usepackage{makecell}
\usepackage{colortbl}

\usetikzlibrary{arrows,positioning}
\usetikzlibrary{plotmarks}
\usetikzlibrary{calc}

\newcommand{\resBlocks}{\bm{N_{rb}}}
\newcommand{\upBlocks}{\bm{N_{up}}}

\newcommand{\TrainingSet}{\bm{T}}
\newcommand{\ValidationSet}{\bm{V}}
\newcommand{\TestSet}{\bm{\Psi}}

\newcommand{\nn}{NN}
\newcommand{\bilin}{Bilinear}
\newcommand{\bicub}{Bicubic}
\newcommand{\lanczos}{Lanczos}
\newcommand{\lanczosb}{Lanczos-B}
\newcommand{\lanczosn}{Lanczos-N}
\newcommand{\lanczosbn}{Lanczos-BN}
\newcommand{\mixed}{Mixed}

\makeatletter
\def\@name{ \emph{Michal Kawulok$^{1,2}$, Szymon Piechaczek$^{1}$, Krzysztof Hrynczenko$^{1}$},  \\ \emph{Pawel Benecki$^{1,2}$, Daniel Kostrzewa$^{1,2}$,} and \emph{Jakub Nalepa$^{1,2}$}\thanks{This work is funded by European Space Agency (SuperDeep project). The work was partially supported by Institute of Informatics funds no. 02/020/BK\_18/0128 (MK) and  BKM-556/RAU2/2018 (PB, DK, JN).} \\}
\makeatother

\title{On training deep networks for satellite image super-resolution}

\address{$^1$Future Processing, Bojkowska 37A, Gliwice, Poland\\$^2$Silesian University of Technology, Institute of Informatics, Akademicka 16, Gliwice, Poland\\\texttt{michal.kawulok@ieee.org}}
\begin{document}
%
\maketitle
\begin{abstract}
The capabilities of super-resolution reconstruction (SRR)---techniques for enhancing image spatial resolution---have been recently improved significantly by the use of deep convolutional neural networks. Commonly, such networks are learned using huge training sets composed of original images alongside their low-resolution counterparts, obtained with bicubic downsampling. In this paper, we investigate how the SRR performance is influenced by the way such low-resolution training data are obtained, which has not been explored up to date. Our extensive experimental study indicates that the training data characteristics have a large impact on the reconstruction accuracy, and the widely-adopted approach is not the most effective for dealing with satellite images. Overall, we argue that developing better training data preparation routines may be pivotal in making SRR suitable for real-world applications.
\end{abstract}
\section{Introduction}
\label{sec:introduction}

\emph{Super-resolution reconstruction} (SRR) is aimed at generating a high-resolution (HR) image from a low-resolution (LR) observation (a single image or multiple images)~\cite{Nasrollahi2014}. SRR is a deeply explored research topic of considerable practical potential, as developing effective SRR techniques may allow for overcoming the spatial resolution limitations of the imaging sensors, which is a common problem in remote sensing.

\subsection{Related work}
Existing single-image SRR methods can be categorized into: (i)~frequency-domain techniques~\cite{Demirel2011}, (ii)~reconstruction-based methods which exploit prior knowledge on the object appearance~\cite{Sun2008}, and (iii)~algorithms that learn the mapping between LR and HR~\cite{Yang2010}. Recently, we have witnessed a breakthrough in the learning-based single-image SRR, attributed to the use of deep \emph{convolutional neural networks} (CNNs).
Deep learning SRR originates from sparse coding~\cite{Yang2010}, aimed at creating a dictionary of LR patches, associated with their HR counterparts. The reconstruction consists in exploiting that dictionary for converting each LR patch from the input image into HR. 

\emph{Super-resolution CNN} (SRCNN)~\cite{Dong2016}, followed by its faster version (FSRCNN)~\cite{Dong2016a}, was proposed for learning the LR-to-HR mapping from a number of LR--HR image pairs. Despite relatively simple architecture, SRCNN outperforms the state-of-the-art example-based methods. In~\cite{Liebel2016}, SRCNN was trained with Sentinel-2 images, which according to the authors improved its capacities of enhancing satellite data. 
Certain limitations of SRCNN were addressed with a \emph{very deep super-resolution} network~\cite{Kim2016}, trained relying on fast residual learning. The domain expertise was exploited in a \emph{sparse coding network}~\cite{Liu2016}, achieving high training speed and model compactness. Recently, \emph{generative adversarial networks} are being actively explored for SRR~\cite{Ledig2017}. They are composed of a \emph{generator} (ResNet in~\cite{Ledig2017}), trained to perform SRR, and a \emph{discriminator} which tries to distinguish the ResNet reconstruction outcomes from real HR images.

\subsection{Contribution}
Deep CNNs for SRR are trained from a dataset of corresponding LR--HR patches. As deep networks commonly require huge amounts of training data, LR images are obtained by subjecting the original HR images to a degradation procedure based on an assumed imaging model. In most works~\cite{Dong2016,Ledig2017,Liebel2016}, bicubic downsampling is applied to transform HR into LR, and in some cases~\cite{Dong2016a}, the \emph{training set} ($\TrainingSet$) is additionally augmented with translation, rotation, and scaling. However, it has not been analyzed whether and how $\TrainingSet$ (including the degradation procedure) influences the reconstruction accuracy.

In this paper, our contribution consists in addressing the aforementioned research gap. 
We investigate the influence of $\TrainingSet$, used for training a CNN, on the reconstruction performance. We trained two different CNNs (Section~\ref{sec:method}) with natural images from the DIV2K single-image SRR benchmark, and with Sentinel-2 images. The trained CNNs are tested in two settings: for reconstructing artificially-degraded satellite images (original images are treated as reference HR data), as well as in a real-world scenario---for original Sentinel-2 images, matched with SPOT and Digital Globe WorldView-4 images of the same region. The results of our extensive experiments (reported in Section~\ref{sec:experiments}) indicate that the degradation procedure used for creating $\TrainingSet$ plays a pivotal role here. Not only does it have a larger impact on the SRR performance than the domain of images exploited for training (natural vs. satellite), but it is also more important than the choice of the CNN architecture.

\section{Deep learning for super-resolution}
\label{sec:method}
In this work, we exploit two CNNs of different complexity, namely: FSRCNN~\cite{Dong2016a}, which is a relatively shallow CNN, and a much deeper residual network (SRResNet~\cite{Ledig2017}), to investigate their behavior in different training scenarios.

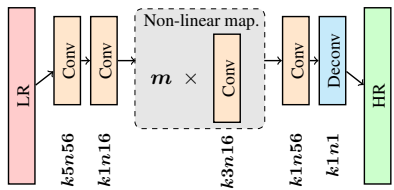
\begin{figure}[ht!]
        \centering
        \resizebox{0.6\columnwidth}{!}{%
            \begin{tikzpicture}[scale=0.8,
                    input/.style={rectangle,draw=black,fill=red!20,inner sep=5pt,minimum height=100pt,minimum width=10pt,text width=5pt,text badly centered,thick},
                    output/.style={rectangle,draw=black,fill=green!20,inner sep=5pt,minimum height=100pt,minimum width=10pt,text width=5pt,text badly centered,thick},
                    convolution/.style={rectangle,draw=black,fill=orange!20,inner sep=5pt,minimum height=50pt,minimum width=10pt,text width=5pt,text badly centered,thick},
                    deconvolution/.style={rectangle,draw=black,fill=cyan!20,inner sep=5pt,minimum height=50pt,minimum width=10pt,text width=5pt,text badly centered,thick},
                    myarrow/.style={thick},
                    dottedarrow/.style={dotted, thick}]

                \newcommand{\lrImg}{{\rotatebox{90}{{LR}}}}
                \newcommand{\srImg}{{\rotatebox{90}{{HR}}}}
                \newcommand{\sep}{5}
                \newcommand{\bigSep}{20}
                \newcommand{\convolution}{{\rotatebox{90}{{Conv}}}}
                \newcommand{\deconvolution}{{\rotatebox{90}{{Deconv}}}}
                \newcommand{\convDescrSep}{15}
                \mathchardef\mhyphen="2D 

                \node (in) [input] {$\lrImg$};
                \coordinate [above=20 pt of in.east](supportPoint0);
                \node (conv1) [convolution, right=10 pt of supportPoint0]{\convolution};
                \node (convDescription) at ($(conv1.south) - (0 pt, \convDescrSep pt)$)[anchor=north] {\rotatebox{90}{$\bm{k5n56}$}};
                \node (conv2) [convolution, right=\sep pt of conv1]{\convolution};
                \node (convDescription) at ($(conv2.south) - (0 pt, \convDescrSep pt)$)[anchor=north] {\rotatebox{90}{$\bm{k1n16}$}};
                \draw [dashed, fill=gray!20, rounded corners] ($(conv2.north) + (22 pt, 5 pt)$) rectangle ++(92 pt, -85 pt);
                \coordinate [below=10 pt of conv2.east](supportPoint3);
                \node (conv3) [convolution, right=54 pt of supportPoint3]{\convolution};
                \node (convDescription) at ($(conv3.south) - (0 pt, 4 pt)$)[anchor=north] {\rotatebox{90}{$\bm{k3n16}$}};
                \node (m) at ($(conv3) - (37 pt, 0 pt)$) {\large{$\bm{m}\enspace\times$}};
                \node (nmDescription) at ($(conv3) - (18 pt, -40 pt)$) {Non-linear map.};
                \coordinate [right=10pt of conv2.east](supportPoint1);
                \coordinate [right=73pt of supportPoint1](supportPoint2);
                \node (conv4) [convolution, right=10 pt of supportPoint2]{\convolution};
                \node (convDescription) at ($(conv4.south) - (0 pt, \convDescrSep pt)$)[anchor=north] {\rotatebox{90}{$\bm{k1n56}$}};
                \node (deconv1) [deconvolution, right=\sep pt of conv4]{\deconvolution};
                \node (convDescription) at ($(deconv1.south) - (0 pt, \convDescrSep pt)$)[anchor=north] {\rotatebox{90}{$\bm{k1n1}$}};
                \node (out) [output, right=186 pt of in]{\srImg};

                \draw[->,myarrow] (in) -- (conv1);
                \draw[->,myarrow] (conv1) -- (conv2);
                \draw[->,myarrow] (conv2) -- (supportPoint1);
                \draw[->,myarrow] (supportPoint2) -- (conv4);
                \draw[->,myarrow] (conv4) -- (deconv1);
                \draw[->,myarrow] (deconv1) -- (out);

            \end{tikzpicture}
        } \caption{FSRCNN architecture for SRR proposed in~\cite{Dong2016a}. For each layer, we report the number of filters $n$, and the size of a square filter $k$ (e.g.,~$k5n56$ means $56$ filters of size $5\times 5$).} \label{fig:fsrcnn}
    \end{figure}

Figure~\ref{fig:fsrcnn} shows the architecture of FSRCNN~\cite{Dong2016a}---the network is composed of five major parts aimed at: (i)~\emph{feature extraction}, realized by the first convolutional layer (denoted as Conv) with $n=56$ kernels of size $k=5\times 5$, (ii)~\emph{shrinking}, performed using $n=16$ kernels ($1\times 1$) to reduce the number of features (from $56$ to $16$), (iii)~\emph{non-linear mapping} using multiple ($m=4$) convolutional layers with $n=16$ kernels ($3\times 3$), (iv)~\emph{expansion} which inverses the shrinking and increases the dimensionality of the feature vectors from $16$ back to $56$, and (v)~\emph{deconvolution} which produces the reconstructed HR image. FSRCNN can be trained faster than SRCNN and it offers real-time performance after training~\cite{Dong2016,Dong2016a}.

\begin{figure}[t!]
        \centering
        \resizebox{1\columnwidth}{!}{%
            \begin{tikzpicture}[scale=0.8,
                    input/.style={rectangle,draw=black,fill=red!20,inner sep=5pt,minimum height=100pt,minimum width=10pt,text width=5pt,text badly centered,thick},
                    output/.style={rectangle,draw=black,fill=green!20,inner sep=5pt,minimum height=100pt,minimum width=10pt,text width=5pt,text badly centered,thick},
                    convolution/.style={rectangle,draw=black,fill=orange!20,inner sep=5pt,minimum height=50pt,minimum width=10pt,text width=5pt,text badly centered,thick},
                    batchNorm/.style={rectangle,draw=black,fill=blue!20,inner sep=5pt,minimum height=50pt,minimum width=10pt,text width=5pt,text badly centered,thick},
                    sum/.style={circle,draw=black,fill=yellow!20,inner sep=2pt,text width=9pt,thick},
                    pShuffle/.style={rectangle,draw=black,fill=cyan!20,inner sep=5pt,minimum height=50pt,minimum width=10pt,text width=5pt,text badly centered,thick},
                    resBlock/.style={rectangle,draw=black,fill=gray!20,inner sep=5pt,minimum height=85pt,minimum width=10pt,text width=5pt,text badly centered,rounded corners},
                    upBlock/.style={rectangle,draw=black,fill=teal!20,inner sep=5pt,minimum height=85pt,minimum width=10pt,text width=5pt,text badly centered,rounded corners},
                    myarrow/.style={thick},
                    dottedarrow/.style={dotted, thick}]

                \newcommand{\lrImg}{{\rotatebox{90}{{LR}}}}
                \newcommand{\srImg}{{\rotatebox{90}{{HR}}}}
                \newcommand{\sep}{5}
                \newcommand{\convDescrSep}{3}
                \newcommand{\bigSep}{20}
                \newcommand{\convolution}{{\rotatebox{90}{{Conv}}}}
                \newcommand{\batchNorm}{{\rotatebox{90}{{BN}}}}
                \newcommand{\pixelShuffle}{{\rotatebox{90}{{PS}}}}
                \newcommand{\elSum}{{\rotatebox{90}{{\Large$+$}}}}
                \newcommand{\nResBlocks}{{\rotatebox{90}{\resBlocks}}}
                \newcommand{\nUpBlocks}{{\rotatebox{90}{\upBlocks}}}

                \node (in) [input] {$\lrImg$};
                \node (conv0) [convolution, right=\sep pt of in]{\convolution};
                \node (convDescription) at ($(conv0.south) - (0 pt, \convDescrSep pt)$)[anchor=north] {\rotatebox{90}{$\bm{k9n64}$}};
                \draw [dashed, fill=gray!20, rounded corners] ($(conv0.north) + (22 pt, 20 pt)$) rectangle ++(155 pt, -139 pt);
                \node (conv1) [convolution, right=\bigSep pt of conv0]{\convolution};
                \node (convDescription) at ($(conv1.south) - (0 pt, \convDescrSep pt)$)[anchor=north] {\rotatebox{90}{$\bm{k3n64}$}};
                \node (bn1) [batchNorm, right=\sep pt of conv1]{\batchNorm};
                \node (conv2) [convolution, right=\sep pt of bn1]{\convolution};
                \node (convDescription) at ($(conv2.south) - (0 pt, \convDescrSep pt)$)[anchor=north] {\rotatebox{90}{$\bm{k3n64}$}};
                \node (bn2) [batchNorm, right=\sep pt of conv2]{\batchNorm};
                \node (sm1) [sum, right=10 pt of bn2]{\elSum};
                \node (resBlockDescription) at ($(conv2) - (0 pt, -35 pt)$) [anchor=south] {$\bm{N_{rb}\,\times\,}$\textbf{RB}};
                \coordinate [right=15pt of conv0](supportPoint1);
                \coordinate [below=65pt of supportPoint1](supportPoint2);
                \coordinate [below=55pt of sm1](supportPoint3);

                \node (conv3) [convolution, right=15 pt of sm1.east]{\convolution};
                \node (convDescription) at ($(conv3.south) - (0 pt, \convDescrSep pt)$)[anchor=north] {\rotatebox{90}{$\bm{k3n64}$}};
                \node (bn3) [batchNorm, right=\sep pt of conv3]{\batchNorm};
                \node (sm2) [sum, right=10 pt of bn3]{\elSum};

                \draw [dashed, fill=gray!20, rounded corners] ($(sm2.north) + (24 pt, 40 pt)$) rectangle ++(70 pt, -139 pt);
                \node (conv4) [convolution, right=\bigSep pt of sm2]{\convolution};
                \node (convDescription) at ($(conv4.south) - (0 pt, \convDescrSep pt)$)[anchor=north] {\rotatebox{90}{$\bm{k3n256}$}};
                \node (ps1) [pShuffle, right=\sep pt of conv4]{\pixelShuffle};
                \node (BlockDescription) at ($(ps1) - (15 pt, -35 pt)$) [anchor=south] {$\bm{N_{ub}\,\times\,}$\textbf{UB}};

                \node (conv5) [convolution, right=\bigSep pt of ps1.east]{\convolution};
                \node (convDescription) at ($(conv5.south) - (0 pt, \convDescrSep pt)$)[anchor=north] {\rotatebox{90}{$\bm{k9n1}$}};
                \node (out) [output, right=10 pt of conv5]{\srImg};

                \coordinate [right=5pt of conv0](supportPoint4);
                \coordinate [below=75pt of supportPoint4](supportPoint5);
                \coordinate [below=65pt of sm2](supportPoint6);

                \draw[->,myarrow] (in) -- (conv0);
                \draw[->,myarrow] (conv0) -- (conv1);
                \draw[->,myarrow] (conv1) -- (bn1);
                \draw[->,myarrow] (bn1) -- (conv2);
                \draw[->,myarrow] (conv2) -- (bn2);
                \draw[->,myarrow] (bn2) -- (sm1);

                \draw[-,myarrow] (supportPoint1) -- (supportPoint2);
                \draw[-,myarrow] (supportPoint2) -- (supportPoint3);
                \draw[->,myarrow] (supportPoint3) -- (sm1.south);

                \draw[->,myarrow] (sm1) -- (conv3);
                \draw[->,myarrow] (conv3) -- (bn3);
                \draw[->,myarrow] (bn3) -- (sm2);

                \draw[->,myarrow] (sm2) -- (conv4);
                \draw[->,myarrow] (conv4) -- (ps1);

                \draw[->,myarrow] (ps1) -- (conv5);

                \draw[->,myarrow] (conv5) -- (out);
                \draw[-,myarrow] (supportPoint4) -- (supportPoint5);
                \draw[-,myarrow] (supportPoint5) -- (supportPoint6);
                \draw[->,myarrow] (supportPoint6) -- (sm2.south);
            \end{tikzpicture}
        } \caption{SRResNet architecture proposed in~\cite{Ledig2017}. In this work, we used $N_{rb}=16$ RBs and $N_{ub}=1$ UB.} \label{fig:resnet}
    \end{figure}
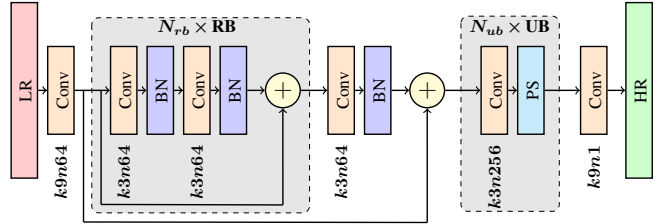

The SRResNet~\cite{Ledig2017} architecture (Fig.~\ref{fig:resnet}) benefits from the residual connections between the layers~\cite{He2016}. 
The \emph{residual blocks} (RBs) are the groups of layers stacked together with the input of the block added to the output of the final layer contained in this block. In SRResNet, each block encompasses two convolutional layers, each followed by a \emph{batch normalization} (BN) layer that neutralizes the internal co-variate shift. The \emph{upsampling blocks} (UBs) allow for image enlargement by \emph{pixel shuffling} (PS) layers that increase the resolution of the features. The number of both RBs and UBs is variable---by increasing the number of RBs, the network may model a better mapping, whereas by changing the number of UBs, we may tune its scaling factor. However, by adding more blocks, the architecture of the network becomes increasingly complex, which makes it harder to train.

\section{Experimental study}
\label{sec:experiments}

\begin{figure}[!h]
\centering
\scriptsize
\newcommand{\w}{0.312}
\newcommand{\smallw}{0.17}
\newcommand{\pos}{0.08,-0.14}
\newcommand{\scale}{0.225}
\newcommand{\textpos}{(main.north west)}
\newcommand{\mytextfont}{\textbf}
\newcommand{\mytextfontinside}{\textbf}

\newcommand{\imagetext}{}

\newcommand{\file}{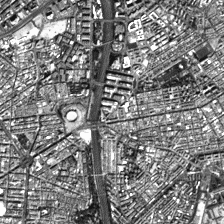}
\resizebox{\columnwidth}{!}{

\begin{tikzpicture}[y=.018\columnwidth, x=.018\columnwidth,
mytext/.style={text=white, font=\scriptsize, anchor=north west}]

    \node (main) [inner sep=0] {\includegraphics[width=\smallw\columnwidth]{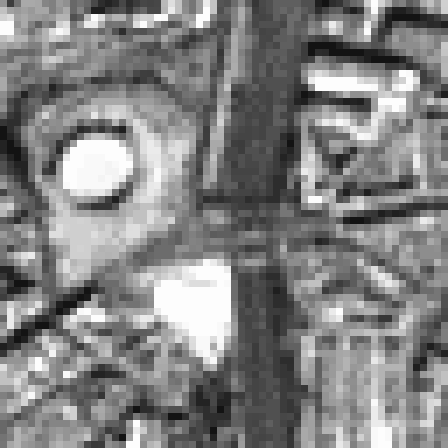}};

    \renewcommand{\imagetext}{{\Large(a)}};
    \node (tmp_text) at \textpos [mytext, text=black] {\mytextfont{\imagetext}};
     \node at ($ (tmp_text.north west) + (0.3, 0.3) $) [mytext,text=black] {\mytextfont{\imagetext}};
     \node at ($ (tmp_text.north west) + (0, 0.3) $) [mytext,text=black] {\mytextfont{\imagetext}};
     \node at ($ (tmp_text.north west) + (0.3, 0) $) [mytext,text=black] {\mytextfont{\imagetext}};
     \node at ($ (tmp_text.north west) + (0.15, 0.15) $) [mytext] {\mytextfontinside{\imagetext}};

    \node (main) at ($ (main.north east) + (0.3,0) $) [anchor=north west, inner sep=0] {\includegraphics[width=\smallw\columnwidth]{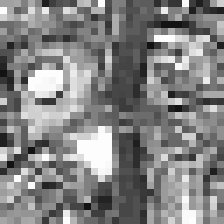}};
    \renewcommand{\imagetext}{{\Large(b)}};
    \node (tmp_text) at \textpos [mytext, text=black] {\mytextfont{\imagetext}};
     \node at ($ (tmp_text.north west) + (0.3, 0.3) $) [mytext,text=black] {\mytextfont{\imagetext}};
     \node at ($ (tmp_text.north west) + (0, 0.3) $) [mytext,text=black] {\mytextfont{\imagetext}};
     \node at ($ (tmp_text.north west) + (0.3, 0) $) [mytext,text=black] {\mytextfont{\imagetext}};
     \node at ($ (tmp_text.north west) + (0.15, 0.15) $) [mytext] {\mytextfontinside{\imagetext}};

    \node (main) at ($ (main.north east) + (0.3,0) $) [anchor=north west, inner sep=0] {\includegraphics[width=\smallw\columnwidth]{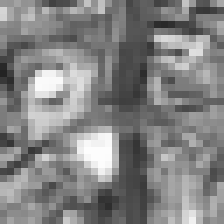}};
    \renewcommand{\imagetext}{{\Large(c)}};
    \node (tmp_text) at \textpos [mytext, text=black] {\mytextfont{\imagetext}};
     \node at ($ (tmp_text.north west) + (0.3, 0.3) $) [mytext,text=black] {\mytextfont{\imagetext}};
     \node at ($ (tmp_text.north west) + (0, 0.3) $) [mytext,text=black] {\mytextfont{\imagetext}};
     \node at ($ (tmp_text.north west) + (0.3, 0) $) [mytext,text=black] {\mytextfont{\imagetext}};
     \node at ($ (tmp_text.north west) + (0.15, 0.15) $) [mytext] {\mytextfontinside{\imagetext}};

    \node (main) at ($ (main.north east) + (0.3,0) $) [anchor=north west, inner sep=0] {\includegraphics[width=\smallw\columnwidth]{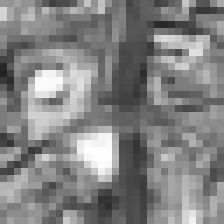}};
    \renewcommand{\imagetext}{{\Large(d)}};
    \node (tmp_text) at \textpos [mytext, text=black] {\mytextfont{\imagetext}};
     \node at ($ (tmp_text.north west) + (0.3, 0.3) $) [mytext,text=black] {\mytextfont{\imagetext}};
     \node at ($ (tmp_text.north west) + (0, 0.3) $) [mytext,text=black] {\mytextfont{\imagetext}};
     \node at ($ (tmp_text.north west) + (0.3, 0) $) [mytext,text=black] {\mytextfont{\imagetext}};
     \node at ($ (tmp_text.north west) + (0.15, 0.15) $) [mytext] {\mytextfontinside{\imagetext}};

    \node (main) at ($ (main.north east) + (0.3,0) $) [anchor=north west, inner sep=0] {\includegraphics[width=\smallw\columnwidth]{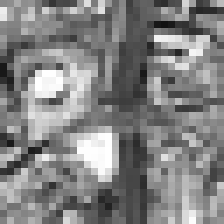}};
    \renewcommand{\imagetext}{{\Large(e)}};
    \node (tmp_text) at \textpos [mytext, text=black] {\mytextfont{\imagetext}};
     \node at ($ (tmp_text.north west) + (0.3, 0.3) $) [mytext,text=black] {\mytextfont{\imagetext}};
     \node at ($ (tmp_text.north west) + (0, 0.3) $) [mytext,text=black] {\mytextfont{\imagetext}};
     \node at ($ (tmp_text.north west) + (0.3, 0) $) [mytext,text=black] {\mytextfont{\imagetext}};
     \node at ($ (tmp_text.north west) + (0.15, 0.15) $) [mytext] {\mytextfontinside{\imagetext}};

    \node (main) at ($ (main.north east) + (0.3,0) $) [anchor=north west, inner sep=0] {\includegraphics[width=\smallw\columnwidth]{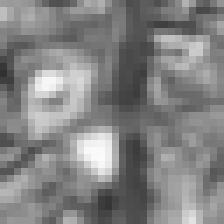}};
    \renewcommand{\imagetext}{{\Large(f)}};
    \node (tmp_text) at \textpos [mytext, text=black] {\mytextfont{\imagetext}};
     \node at ($ (tmp_text.north west) + (0.3, 0.3) $) [mytext,text=black] {\mytextfont{\imagetext}};
     \node at ($ (tmp_text.north west) + (0, 0.3) $) [mytext,text=black] {\mytextfont{\imagetext}};
     \node at ($ (tmp_text.north west) + (0.3, 0) $) [mytext,text=black] {\mytextfont{\imagetext}};
     \node at ($ (tmp_text.north west) + (0.15, 0.15) $) [mytext] {\mytextfontinside{\imagetext}};

    \node (main) at ($ (main.north east) + (0.3,0) $) [anchor=north west, inner sep=0] {\includegraphics[width=\smallw\columnwidth]{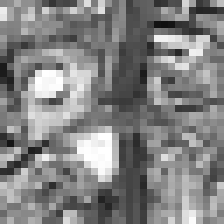}};
    \renewcommand{\imagetext}{{\Large(g)}};
    \node (tmp_text) at \textpos [mytext, text=black] {\mytextfont{\imagetext}};
     \node at ($ (tmp_text.north west) + (0.3, 0.3) $) [mytext,text=black] {\mytextfont{\imagetext}};
     \node at ($ (tmp_text.north west) + (0, 0.3) $) [mytext,text=black] {\mytextfont{\imagetext}};
     \node at ($ (tmp_text.north west) + (0.3, 0) $) [mytext,text=black] {\mytextfont{\imagetext}};
     \node at ($ (tmp_text.north west) + (0.15, 0.15) $) [mytext] {\mytextfontinside{\imagetext}};

    \node (main) at ($ (main.north east) + (0.3,0) $) [anchor=north west, inner sep=0] {\includegraphics[width=\smallw\columnwidth]{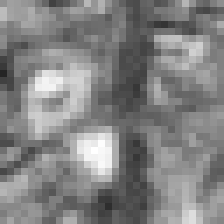}};

    \renewcommand{\imagetext}{{\Large(h)}};
    \node (tmp_text) at \textpos [mytext, text=black] {\mytextfont{\imagetext}};
     \node at ($ (tmp_text.north west) + (0.3, 0.3) $) [mytext,text=black] {\mytextfont{\imagetext}};
     \node at ($ (tmp_text.north west) + (0, 0.3) $) [mytext,text=black] {\mytextfont{\imagetext}};
     \node at ($ (tmp_text.north west) + (0.3, 0) $) [mytext,text=black] {\mytextfont{\imagetext}};
     \node at ($ (tmp_text.north west) + (0.15, 0.15) $) [mytext] {\mytextfontinside{\imagetext}};

\end{tikzpicture}
}
  \caption{A $64 \times 64$ patch (a) degraded with different techniques: b)~NN, c)~bilinear, d)~bicubic, e)~Lanczos, f)~Lanczos-B, g)~Lanczos-N, and h)~Lanczos-BN.}
  \label{fig:train_examples}
\end{figure}
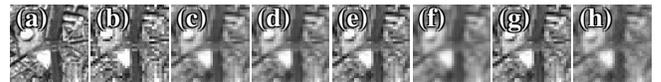

We trained FSRCNN and SRResNet using natural images from the DIV2K dataset\footnote{Available at \scriptsize \url{https://data.vision.ee.ethz.ch/cvl/DIV2K}}, and Sentinel-2 images. From these images, the patches were extracted randomly to create $\TrainingSet$ and \emph{validation set} ($\ValidationSet$), as specified in Table~\ref{tab:data}. LR images were obtained from HR ones using different downsampling techniques: nearest neighbor (NN), bilinear, bicubic, and Lanczos. We also created a mixed set---the downsampling technique was randomly selected for each image. For Lanczos, we additionally applied Gaussian blur with $\sigma_b=0.7$ (Lanczos-B), Gaussian noise with $\sigma_n=0.01$ (Lanczos-N), and both blur and noise with $\sigma_n=0.022$ (Lanczos-BN). Examples of patches in $\TrainingSet$ are shown in Fig.~\ref{fig:train_examples}. We used Python with Keras to implement the CNNs. The experiments were run on an Intel i9 4 GHz computer with 64 GB RAM, and two RTX 2080 8 GB GPUs. We used ADAM optimizer with learning rate of $10^{-3}$. The optimization stops, if after 50 epochs the accuracy over $\ValidationSet$ does not increase.
\begin{table}[!t]
\centering
\caption{Datasets used for training FSRCNN and SRResNet.}\label{tab:data}
\scriptsize
\renewcommand{\tabcolsep}{1.1mm}
\begin{tabular}{rcccc}
\Xhline{2\arrayrulewidth}
Dataset & No. of patches in $\TrainingSet$ & No. of patches in $\ValidationSet$ & LR patch size & HR patch size \\ \hline
DIV2K & 12800 & 1600 & 112$\times$112 & 224$\times$224 \\
Sentinel & 4825 & 535 & 112$\times$112 & 224$\times$224 \\
\Xhline{2\arrayrulewidth}
\end{tabular}
\end{table}

\begin{table*}[!ht]
\centering
\caption{Reconstruction accuracy obtained for $\TestSet$ after training FSRCNN and SRResNet using different $\TrainingSet$'s (best scores for each category are marked as bold). The scenarios commonly reported in the literature are marked as gray.}\label{tab:scores}
\scriptsize
\renewcommand{\tabcolsep}{1.1mm}
\resizebox{\textwidth}{!}{
\begin{tabular}{cr ccccc c ccccc c ccccc c ccccc}
\Xhline{2\arrayrulewidth}

\multicolumn{2}{r}{ $\TestSet \rightarrow$} & \multicolumn{11}{c}{Artificially-degraded (AD) satellite images} & & \multicolumn{11}{c}{Real satellite (RS) images} \\ \cline{3-13} \cline{15-25}
\multicolumn{2}{r}{SRR method $\rightarrow$} & \multicolumn{5}{c}{FSRCNN~\cite{Dong2016a}} & & \multicolumn{5}{c}{SRResNet~\cite{Ledig2017}} & & \multicolumn{5}{c}{FSRCNN~\cite{Dong2016a}} & & \multicolumn{5}{c}{SRResNet~\cite{Ledig2017}} \\ \cline{3-7} \cline{9-13} \cline{15-19} \cline{21-25}

\multicolumn{2}{r}{Downsampling of $\TrainingSet$ $\downarrow$ } & PSNR & SSIM & UIQI & VIF & KFS& & PSNR & SSIM & UIQI & VIF & KFS & & PSNR & SSIM & UIQI & VIF & KFS & & PSNR & SSIM & UIQI & VIF & KFS \\ \hline
\multirow{8}{*}{\begin{sideways}DIV2K\end{sideways}}
 & \nn & \textbf{31.95} & \textbf{0.915} & \textbf{0.891} & \textbf{0.545} & \textbf{12.92} &  & \textbf{30.81} & \textbf{0.905} & \textbf{0.879} & \textbf{0.523} & \textbf{12.708} &  & 16.79 & 0.454 & 0.268 & 0.122 & 2.638 &  & 17.29 & 0.439 & 0.263 & 0.117 & 2.64 \\
 & \bilin & 22.26 & 0.679 & 0.652 & 0.351 & 7.669 &  & 22.03 & 0.67 & 0.643 & 0.35 & 7.747 &  & 16.83 & 0.454 & \textbf{0.292} & 0.125 & 2.768 &  & 16.77 & 0.457 & \textbf{0.292} & 0.124 & 2.762 \\
 & \bicub & \cellcolor{gray!50} 26.61 & \cellcolor{gray!50} 0.819 & 0.792 & 0.435 & 10.209 &  & \cellcolor{gray!50} 26.84 & \cellcolor{gray!50} 0.82 & 0.794 & 0.44 & 10.375 &  & 16.44 & 0.433 & 0.262 & 0.109 & 2.61 &  & 16.97 & 0.465 & 0.287 & 0.124 & 2.705 \\
 & \lanczos & 28 & 0.844 & 0.818 & 0.454 & 11.093 &  & 28.57 & 0.854 & 0.829 & 0.466 & 11.364 &  & 16.9 & \textbf{0.459} & 0.283 & \textbf{0.126} & 2.664 &  & 16.62 & 0.47 & 0.274 & 0.117 & 2.661 \\
 & \lanczosb & 11.02 & 0.175 & 0.168 & 0.118 & 3.591 &  & 10.98 & 0.167 & 0.162 & 0.11 & 3.514 &  & 15.32 & 0.313 & 0.21 & 0.098 & \textbf{2.817} &  & 15.45 & 0.336 & 0.218 & 0.099 & \textbf{2.819} \\
 & \lanczosn & 28.6 & 0.866 & 0.836 & 0.473 & 11.429 &  & 28.14 & 0.869 & 0.836 & 0.47 & 11.256 &  & \textbf{16.91} & 0.456 & 0.271 & 0.124 & 2.624 &  & \textbf{17.74} & \textbf{0.479} & 0.271 & 0.122 & 2.634 \\
 & \lanczosbn & 19.97 & 0.676 & 0.624 & 0.337 & 5.94 &  & 18.35 & 0.583 & 0.543 & 0.299 & 5.451 &  & 16.49 & 0.434 & 0.257 & 0.117 & 2.659 &  & 16.47 & 0.46 & 0.263 & 0.118 & 2.689 \\
 & \mixed & 30.16 & 0.885 & 0.858 & 0.481 & 11.504 &  & 28.5 & 0.856 & 0.829 & 0.456 & 11.729 &  & 16.84 & 0.453 & 0.289 & \textbf{0.126} & 2.717 &  & 16.32 & 0.476 & 0.29 & \textbf{0.126} & 2.737 \\
\hline \multirow{8}{*}{\begin{sideways}Sentinel-2\end{sideways}}
 & \nn & \textbf{31.64} & \textbf{0.91} & \textbf{0.88} & \textbf{0.531} & \textbf{12.794} &  & \textbf{31.59} & \textbf{0.908} & \textbf{0.875} & \textbf{0.527} & \textbf{12.43} &  & 16.88 & 0.438 & 0.242 & 0.11 & 2.553 &  & 16.08 & 0.441 & 0.254 & 0.11 & 2.591 \\
 & \bilin & 23.01 & 0.701 & 0.676 & 0.358 & 7.467 &  & 23.01 & 0.669 & 0.632 & 0.308 & 6.378 &  & \textbf{17.12} & 0.491 & \textbf{0.292} & 0.124 & 2.772 &  & 16.9 & 0.507 & \textbf{0.279} & 0.109 & 2.682 \\
 & \bicub & \cellcolor{gray!50} 27.82 & \cellcolor{gray!50} 0.837 & 0.804 & 0.426 & 10.636 &  & \cellcolor{gray!50} 27.97 & \cellcolor{gray!50} 0.844 & 0.797 & 0.435 & 9.702 &  & 16.38 & \textbf{0.502} & 0.287 & 0.126 & 2.769 &  & 16.93 & 0.458 & 0.227 & 0.079 & 2.568 \\
 & \lanczos & 28.41 & 0.85 & 0.823 & 0.459 & 11.04 &  & 26.18 & 0.833 & 0.807 & 0.445 & 11.04 &  & 16.93 & 0.49 & 0.285 & 0.126 & 2.686 &  & 15.52 & 0.482 & 0.254 & 0.105 & 2.593 \\
 & \lanczosb & 12.2 & 0.216 & 0.207 & 0.134 & 3.722 &  & 12.21 & 0.221 & 0.202 & 0.073 & 2.54 &  & 15.63 & 0.337 & 0.216 & 0.099 & \textbf{2.806} &  & 15.49 & 0.4 & 0.225 & 0.098 & \textbf{2.769} \\
 & \lanczosn & 28.67 & 0.865 & 0.839 & 0.474 & 11.348 &  & 28.68 & 0.868 & 0.842 & 0.48 & 11.557 &  & 16.88 & 0.487 & 0.275 & \textbf{0.127} & 2.652 &  & \textbf{17.35} & \textbf{0.528} & 0.265 & \textbf{0.122} & 2.664 \\
 & \lanczosbn & 20.7 & 0.702 & 0.663 & 0.342 & 6.014 &  & 18.79 & 0.6 & 0.553 & 0.281 & 5.203 &  & 16.53 & 0.455 & 0.269 & 0.12 & 2.718 &  & 17.02 & 0.515 & 0.261 & 0.114 & 2.71 \\
 & \mixed & 28.23 & 0.847 & 0.817 & 0.431 & 10.576 &  & 20.83 & 0.843 & 0.805 & 0.45 & 9.555 &  & 16.99 & 0.461 & 0.291 & 0.126 & 2.778 &  & 13.47 & 0.46 & 0.237 & 0.084 & 2.563 \\

\Xhline{2\arrayrulewidth}
\end{tabular}
}
\end{table*}

After training, the nets were tested using two kinds of test sets ($\TestSet$): (i)~\emph{artificially-degraded} (AD) images---10 HR images of size $500\times 500$ pixels, bicubically downsampled to $250 \times 250$ pixels, and (ii)~\emph{real satellite} (RS) images acquired at different resolution---we used three Sentinel-2 scenes as LR, two of which are matched with SPOT images and one is matched with Digital Globe WorldView-4 image. We evaluate the reconstruction accuracy relying on \emph{peak signal-to-noise ratio} (PSNR), \emph{structural similarity index} (SSIM), \emph{visual information fidelity} (VIF), \emph{universal image quality index} (UIQI), and \emph{keypoint features similarity} (KFS)~\cite{Benecki2018AA}. For all these metrics, higher values indicate higher similarity between the reconstruction outcome and the reference image.

\begin{figure*}[!th]
\centering
\renewcommand{\tabcolsep}{0.2cm}
\scriptsize
\newcommand{\w}{0.312}
\newcommand{\wenlarged}{0.312}
\newcommand{\pos}{0.08,-0.14}
\newcommand{\scale}{0.225}
\newcommand{\textpos}{(main.north west)}
\newcommand{\mytextfont}{\textbf}
\newcommand{\mytextfontinside}{\textbf}

\newcommand{\imagetext}{}

\newcommand{\file}{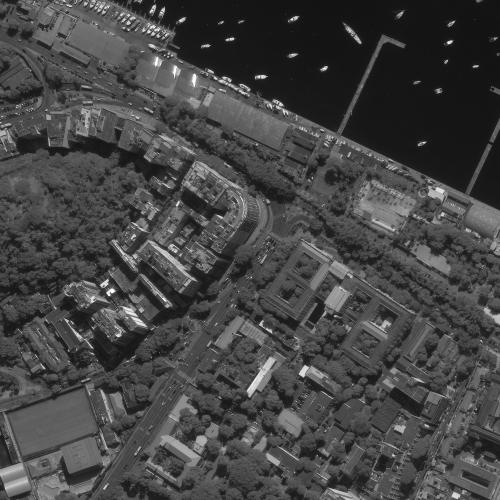}
\newcommand{\zoomfile}{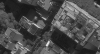}

\resizebox{0.88\textwidth}{!}{


}
  \caption{Examples of reconstructing an artificially degraded Digital Globe WorldView-4 image ($250 \times 250$ pixels, presenting Rio de Janeiro, Brasil) using SRResNet and FSRCNN, trained with $\TrainingSet$'s obtained relying on different degradation procedures.}
  \label{fig:degraded}
\end{figure*}

In Table~\ref{tab:scores}, we show the reconstruction accuracy obtained with FSRCNN and SRResNet trained using different $\TrainingSet$'s. We highlight PSNR and SSIM scores (in gray) for AD after training with bicubically downsampled $\TrainingSet$, as this is the scenario most often reported in the literature. From these scores, SRResNet is slightly better than FSRCNN, and using satellite data for training appears to be beneficial. However, from all the scores, it is clear that the degradation procedure is more significant than both the type of images in $\TrainingSet$ and the network architecture. Actually, the nets trained with $\TrainingSet$'s based on NN perform in the best way, which can also be assessed qualitatively from Fig.~\ref{fig:degraded}. If $\TrainingSet$ is blurred (Lanczos-B), then the image sharpening is too strong, resulting in many high-frequency artifacts. A surprising outcome can be observed for SRResNet (Bicubic, Sentinel)---the details in the sea area are lost after reconstruction, but the land area is reliably restored.

\begin{figure*}[!ht]
\centering
\renewcommand{\tabcolsep}{0.2cm}
\scriptsize
\newcommand{\w}{0.312}
\newcommand{\wenlarged}{0.312}
\newcommand{\pos}{0.148,-0.16}
\newcommand{\scale}{0.2}
\newcommand{\textpos}{(main.north west)}
\newcommand{\mytextfont}{\textbf}
\newcommand{\mytextfontinside}{\textbf}

\newcommand{\imagetext}{}

\newcommand{\file}{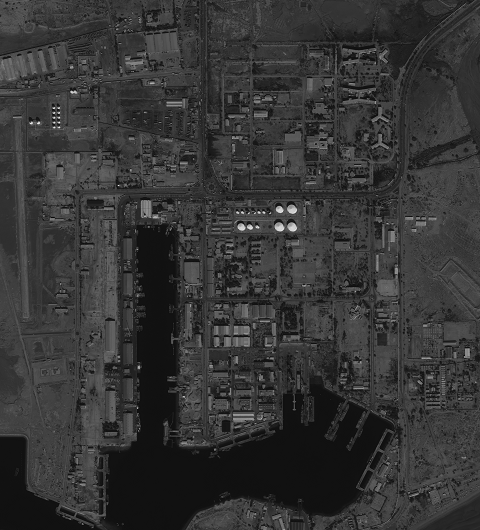}
\newcommand{\zoomfile}{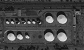}

\resizebox{0.88\textwidth}{!}{


}
  \caption{Examples of reconstructing a real Sentinel-2 image ($240\times 266$ pixels, presenting {Bandar Abbas}, Iran) using SRResNet and FSRCNN, trained with different $\TrainingSet$'s based on different degradation procedures. HR image (SPOT) is given for reference.}
  \label{fig:real}
\end{figure*}

For RS images, it is not clear from the reported metrics (Table~\ref{tab:scores}), which $\TrainingSet$ is the best. The values are much lower than for AD, as the HR images used for reference are acquired using a different sensor, so even if an image is well reconstructed, it is substantially different from HR. From Fig.~\ref{fig:real}, it can be seen that NN downsampling (best for AD) results in a blurry outcome. Interestingly, Lanczos-B (very poor for AD), delivers better results in this case (and it is consistently picked by the KFS metric---the similarity to HR in the domain of the detected keypoints is the highest here). Similarly to AD, severe artifacts in the sea area can be observed for SRResNet trained with some $\TrainingSet$'s (Mixed and NN, for Sentinel). In our opinion, visually most plausible results are obtained using bilinear downsampling (for both Sentinel and DIV2K), which is also reflected with the highest UIQI scores in Table~\ref{tab:scores}.


\section{Conclusions}
\label{sec:conclusions}

In this paper, we reported our experimental study on preparing the data to train deep CNNs for satellite image SRR. The results indicate that the degradation procedure used to generate the training data has a tremendous impact on the SRR performance, which is usually neglected in the literature. Furthermore, it is worth noting that much deeper architecture of SRResNet does not seem to outperform a relatively simple FSRCNN, when appropriate $\TrainingSet$ is used. 

Currently, we are exploring how to combine different degradation procedures, including data augmentation techniques, to create training sets which better reflect the actual imaging conditions. We expect that this will allow deep CNNs to increase their performance for real satellite images.


\begin{thebibliography}{10}
\providecommand{\url}[1]{#1}
\csname url@samestyle\endcsname
\providecommand{\newblock}{\relax}
\providecommand{\bibinfo}[2]{#2}
\providecommand{\BIBentrySTDinterwordspacing}{\spaceskip=0pt\relax}
\providecommand{\BIBentryALTinterwordstretchfactor}{4}
\providecommand{\BIBentryALTinterwordspacing}{\spaceskip=\fontdimen2\font plus
\BIBentryALTinterwordstretchfactor\fontdimen3\font minus
  \fontdimen4\font\relax}
\providecommand{\BIBforeignlanguage}[2]{{%
\expandafter\ifx\csname l@#1\endcsname\relax
\typeout{** WARNING: IEEEtranS.bst: No hyphenation pattern has been}%
\typeout{** loaded for the language `#1'. Using the pattern for}%
\typeout{** the default language instead.}%
\else
\language=\csname l@#1\endcsname
\fi
#2}}
\providecommand{\BIBdecl}{\relax}
\BIBdecl

\bibitem{Benecki2018AA}
P.~Benecki, M.~Kawulok, D.~Kostrzewa, and L.~Skonieczny, ``Evaluating
  super-resolution reconstruction of satellite images,'' \emph{Acta
  Astronautica}, vol. 153, pp. 15--25, 2018.

\bibitem{Demirel2011}
H.~Demirel and G.~Anbarjafari, ``Discrete wavelet transform-based satellite
  image resolution enhancement,'' \emph{IEEE TGRS}, vol.~49, no.~6, pp.
  1997--2004, 2011.

\bibitem{Dong2016}
C.~Dong, C.~C. Loy, K.~He, and X.~Tang, ``Image super-resolution using deep
  convolutional networks,'' \emph{IEEE TPAMI}, vol.~38, no.~2, pp. 295--307,
  2016.

\bibitem{Dong2016a}
C.~Dong, C.~C. Loy, and X.~Tang, ``Accelerating the super-resolution
  convolutional neural network,'' in \emph{Proc. ECCV}.\hskip 1em plus 0.5em
  minus 0.4em\relax Springer, 2016, pp. 391--407.

\bibitem{He2016}
K.~He, X.~Zhang, S.~Ren, and J.~Sun, ``Deep residual learning for image
  recognition,'' in \emph{Proc. IEEE CVPR}, 2016, pp. 770--778.

\bibitem{Kim2016}
J.~Kim, J.~Kwon~Lee, and K.~Mu~Lee, ``Accurate image super-resolution using
  very deep convolutional networks,'' in \emph{Proc. IEEE CVPR}, 2016, pp.
  1646--1654.

\bibitem{Ledig2017}
C.~Ledig, L.~Theis, F.~Husz{\'a}r \emph{et~al.}, ``Photo-realistic single image
  super-resolution using a generative adversarial network.'' in \emph{Proc.
  CVPR}, vol.~2, no.~3, 2017, p.~4.

\bibitem{Liebel2016}
L.~Liebel and M.~K{\"o}rner, ``Single-image super resolution for multispectral
  remote sensing data using {CNNs},'' in \emph{Proc. ISPRSC}, 2016, pp.
  883--890.

\bibitem{Liu2016}
D.~Liu, Z.~Wang, B.~Wen \emph{et~al.}, ``Robust single image super-resolution
  via deep networks with sparse prior,'' \emph{IEEE TIP}, vol.~25, no.~7, pp.
  3194--3207, 2016.

\bibitem{Nasrollahi2014}
K.~Nasrollahi and T.~B. Moeslund, ``Super-resolution: a comprehensive survey,''
  \emph{Machine vision and applications}, vol.~25, no.~6, pp. 1423--1468, 2014.

\bibitem{Sun2008}
J.~Sun, Z.~Xu, and H.-Y. Shum, ``Image super-resolution using gradient profile
  prior,'' in \emph{IEEE CVPR}.\hskip 1em plus 0.5em minus 0.4em\relax IEEE,
  2008, pp. 1--8.

\bibitem{Yang2010}
J.~Yang, J.~Wright, T.~S. Huang, and Y.~Ma, ``Image super-resolution via sparse
  representation,'' \emph{IEEE TIP}, vol.~19, no.~11, pp. 2861--2873, 2010.

\end{thebibliography}

\end{document}